\documentclass[11pt,a4paper]{article}
\usepackage[hyperref]{naaclhlt2018}
\usepackage{times}
\usepackage{latexsym}
\usepackage{graphicx}
\usepackage[space]{grffile}
\usepackage{url}

\usepackage{subcaption}
\aclfinalcopy 

\title{ntuer at SemEval-2019 Task 3:  Emotion Classification with Word and Sentence Representations in RCNN}

\author{Peixiang Zhong, Chunyan Miao\\
  School of Computer Science and Engineering \\
  Nanyang Technological University Singapore \\
  {\tt peixiang001@e.ntu.edu.sg}, {\tt ascymiao@ntu.edu.sg} \\}

\date{}

\begin{document}
\maketitle
\begin{abstract}
In this paper we present our model on the task of emotion detection in textual conversations in SemEval-2019. Our model extends the Recurrent Convolutional Neural Network (RCNN) by using external fine-tuned word representations and DeepMoji sentence representations. We also explored several other competitive pre-trained word and sentence representations including ELMo, BERT and InferSent but found inferior performance. In addition, we conducted extensive sensitivity analysis, which empirically shows that our model is relatively robust to hyper-parameters. Our model requires no handcrafted features or emotion lexicons but achieved good performance with a micro-F1 score of 0.7463. 
\end{abstract}

\section{Introduction}
Emotions are psychological and physiological states generated in humans in reaction to internal or external events. Messages in human conversations inherently convey emotions. With the rise of social media platforms such as Twitter, as well as chatbots such as Amazon Alexa, there is an emerging need for machines to understand human emotions in conversations, which has a wide range of applications such as opinion analysis in customer support \cite{devillers2002annotation} and providing emotion-aware responses \cite{zhong2018affect}. SemEval-2019 Task 3: EmoContext \cite{SemEval2019Task3} is designed to promote research in this task.

This task is to detect emotions in textual conversations. Each conversation is composed of three turns of utterances and the objective is to detect the emotion of the last utterance given the first two utterances as the context. The emotions in this classification task include happy, sad, angry and others, adapted from the well-known Ekman's six basic emotions: anger, disgust, fear, happiness, sadness, and surprise \cite{ekman1992argument}. The evaluation criteria is micro-averaged F1 score since the data is extremely unbalanced, as shown in Table 1.

In recent years, pre-trained word and sentence representations achieved very competitive performance in many NLP tasks, e.g., fine-tuned word embeddings using distant training \cite{cliche2017bb_twtr} and tweet sentence representations DeepMoji \cite{felbo2017using} on sentiment analysis, and contextualized word representations BERT \cite{devlin2018bert} on 11 NLP tasks. Motivated by these successes, in this task we explored different word and sentence representations. We then fed these representations into a Recurrent Convolutional Neural Network (RCNN) \cite{lai2015recurrent} for classification. RCNN includes a Long short-term memory (LSTM) network \cite{hochreiter1997long} to capture word ordering information and a max-pooling layer \cite{scherer2010evaluation} to learn discriminative features. We also experimented LSTM and CNN in our preliminary analysis but achieved worse performance as compared to RCNN. Our final system adopted fine-tuned word embeddings and DeepMoji as our choices of word and sentence representations, respectively, due to their superior performance on the validation dataset. The code is publicly available at Github\footnote{https://github.com/zhongpeixiang/SemEval2019-Task3-EmotionDetection}.


\begin{table*}[t!]
\small
\begin{center}
\begin{tabular}{|c|c|c|c|c|c|c|}
\hline \bf Dataset Split & \bf Size & \bf \#Happy & \bf \#Sad &\bf \#Angry &\bf \#Others &\bf Average Utterance Length\\ \hline
Train & 30160 & 4243 & 5463 & 5506 & 14948 &5.22 \\
\hline
Val & 2755 & 142 & 125 & 150 & 2338 &5.05\\
\hline
Test & 5509 & 284 & 250 & 298 & 4677 &5.05\\
\hline
\end{tabular}
\end{center}
\caption{Total number of conversations and their distributions over each emotion class for each dataset split. Average number of tokens per utterance for each dataset split are also reported.}
\end{table*}

\section{Related Work}
\label{relaed work}
Emotion detection in textual conversations is an under-explored research task. The majority of existing works focused on the multi-modality settings \cite{devillers2002annotation, hazarika2018conversational,majumder2018dialoguernn}. \citet{chatterjee2019understanding} is one of the early works on the textual modality that first collected the dataset used in this task and then proposed an LSTM model with both semantic and sentiment embeddings to classify emotions. This task is also closely related to sentiment analysis \cite{pang2008opinion} where the opinions of a piece of text is to be identified. One major difference between them is that this task detects emotions only on the last portion of a piece of text and the rest is treated as context.

Our model leverages pre-trained word and sentence representations. There is a research trend on word and sentence embeddings after the invention of Word2Vec  \cite{mikolov2013distributed}. \citet{cliche2017bb_twtr} fine-tuned word embeddings using CNN-based sentiment classification model and distant training \cite{go2009twitter}. \citet{peters2018deep} proposed a contextualized word embedding named ELMo to incorporate context information and solve the polysemy issues in conventional word embeddings. \citet{devlin2018bert} proposed another contextualized word embedding named BERT by extending the context to both directions and training on the masked language modelling task. \citet{kiros2015skip} proposed a sentence-level representation named SkipThought, which shares similar ideas to Word2Vec but operates on sentence level. \citet{conneau2017supervised} proposed InferSent by learning sentence representations on natural language inference tasks. \citet{felbo2017using} proposed DeepMoji by learning tweet sentence representations in the emoji classification task using 1246 million tweets and distant training.

Our RCNN model is closely related to neural network based sentiment analysis models. Two of the most popular models are LSTMs and CNNs. LSTM-based models can capture the word ordering information and have achieved the state-of-the-art performance on many sentiment analysis datasets \cite{gray2017gpu, liu2018recurrent, howard2018universal}. CNN-based models can capture local dependencies, discriminative features, and are parallelizable for efficient computation \cite{kim2014convolutional, johnson2017deep}.

\section{System Description}
\label{system description}

\begin{figure*}[!t]
\centering
\includegraphics[scale=0.41]{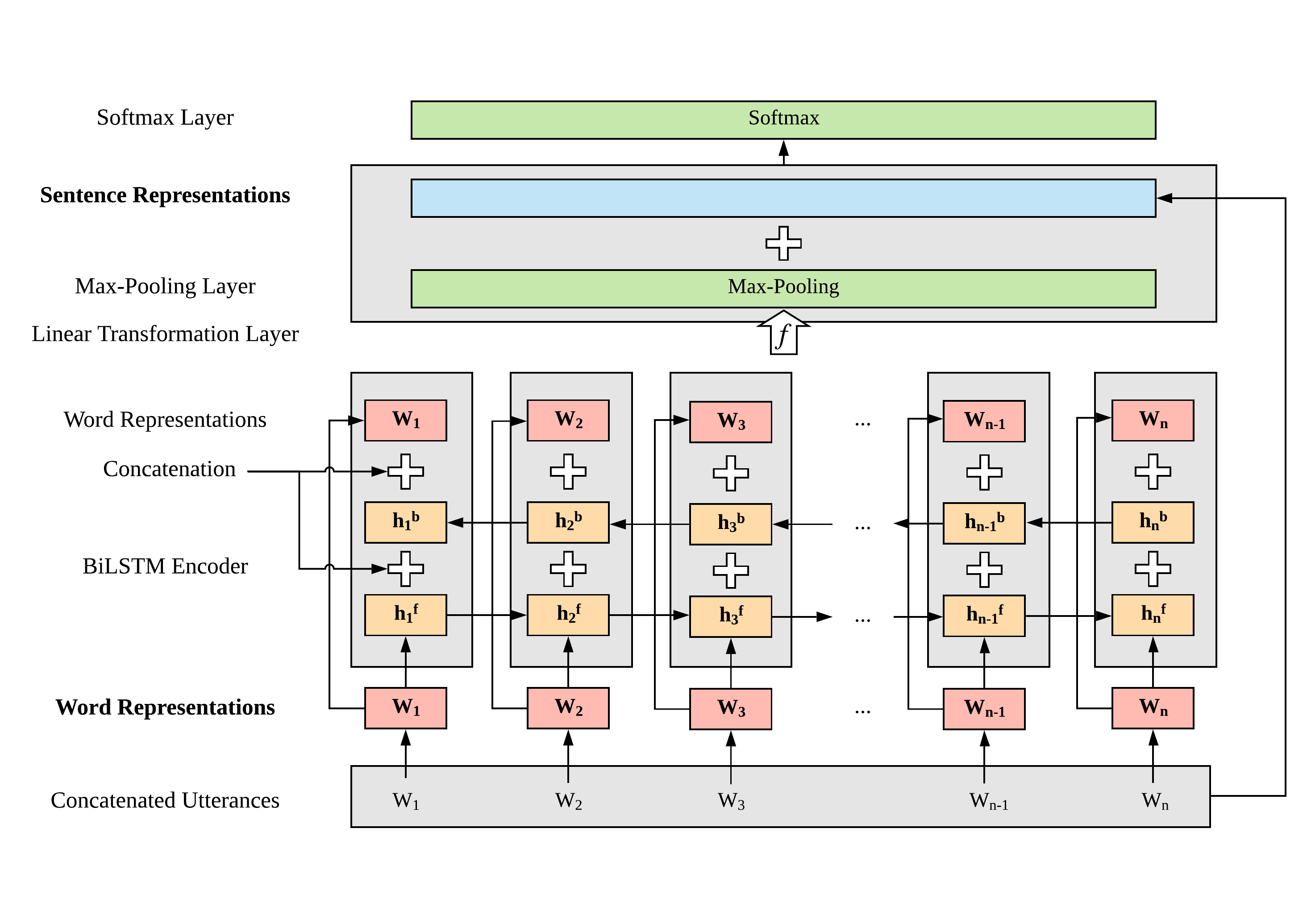}
\caption{Overall architecture of our proposed model}
\label{fig: architecture}
\end{figure*}
In this section we describe our data preprocessing procedures and illustrate how we leverage pre-trained word and sentence representations in our RCNN model. The overall architecture is depicted in Figure \ref{fig: architecture}.

\subsection{Data Preprocessing}
We concatenated three utterances as one sentence, separated by EOS tokens. We used the tokenizer from Spacy\footnote{https://spacy.io/} for tokenization. We removed training sentences that have more than 75 tokens. We removed duplicate punctuations and spaces. We kept all remaining tokens in the training dataset as the vocabulary. 

\subsection{Pre-trained Word Representation}
We fine-tuned the word embeddings obtained from \cite{baziotis2017datastories}, which has an embedding size of 100 and is pre-trained on 330M English Twitter messages using Glove \cite{pennington2014glove}. The fine-tuning is conducted on the binary sentiment classification task using the basic CNN model \cite{kim2014convolutional} on 1.6 million tweets \cite{go2009twitter}. These tweets are labelled with positive and negative sentiments. Fine-tuning on these tweets introduces sentiment-discriminative features to word embeddings \cite{cliche2017bb_twtr}. The CNN model has kernel sizes of 1, 2, and 3. Each kernel size has 300 filters. During fine-tuning, the embedding layer is first frozen for one epoch and then unfrozen for another three epochs.

\subsection{Pre-trained Sentence Representation}
We adopted DeepMoji \cite{felbo2017using} as the sentence representations in our model. Each sentence will be encoded into a vector of size 2304. DeepMoji is trained on the 64-class emoji classification task using 1246 million tweets. Since emoji reflects emotions and sentiments, DeepMoji is an ideal model to provide emotion-discriminative sentence representations. We also explored InferSent \cite{conneau2017supervised}, another sentence representation model with competitive performance on sentence classification and information retrieval tasks \cite{perone2018evaluation}. 

\subsection{RCNN}
As shown in Figure \ref{fig: architecture}, we fed word and sentence representations into a RCNN model. The RCNN model mainly comprises of a two-layer Bi-directional LSTM (BiLSTM), a linear transformation layer and a max-pooling layer. At the embedding layer, each sentence is transformed to a sequence of word embeddings $W_i$ of size 100 using our pre-trained word representations, where $i=1,2,...,n$, and $n$ is the number of tokens in the concatenated utterance. The BiLSTM encodes these word embeddings into hidden states $h_i^f, h_i^b$ in both forward and backward directions, respectively, where each direction has a hidden size of 200. The hidden states in both directions are concatenated together, along with the word representations $W_i$ to form a vector of size 500. A linear transformation is then applied to project the resulted vector into a vector of size 200, followed by a max-pooling layer to extract discriminative sentence features. Finally, the DeepMoji sentence representation is concatenated with the pooled vector to form a final sentence representation of size 2504, followed by a softmax layer for classification. 

\subsection{Training}
We train our model on the training dataset and fine-tune on the validation dataset based on the micro-F1 score. Since the dataset is highly unbalanced, we use weighted cross-entropy loss for optimization, where the weights are the ratio of validation dataset label distribution to training dataset label distribution, followed by a normalization to ensure that the sum of weights is 1. We use Adam \cite{kingma2014adam} optimizer with a learning rate of 0.0005 and batch size of 64. We clip the norm of gradients to 5. We trained our model 6 epochs. The learning rate is annealed by a factor of 0.2 every epoch after epoch 5. We also freeze the embedding for the first two epochs. We use dropout rate of 0.5 in BiLSTM and 0.7 in linear layers. The model is implemented in PyTorch.

\section{Result Analysis}
\label{result analysis}

\begin{figure*}
    \centering
    \begin{subfigure}[b]{0.31\textwidth}
        \centering
        \includegraphics[scale=0.33]{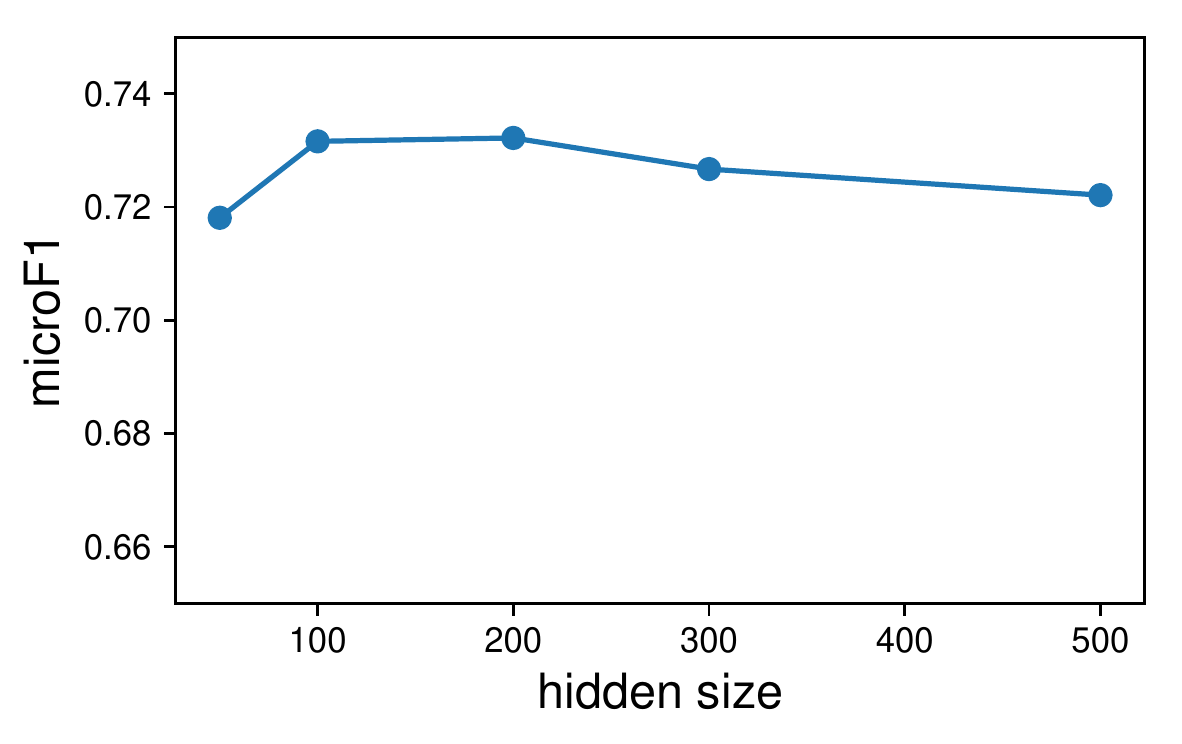}
    \end{subfigure}
    \begin{subfigure}[b]{0.31\textwidth}
        \centering
        \includegraphics[scale=0.33]{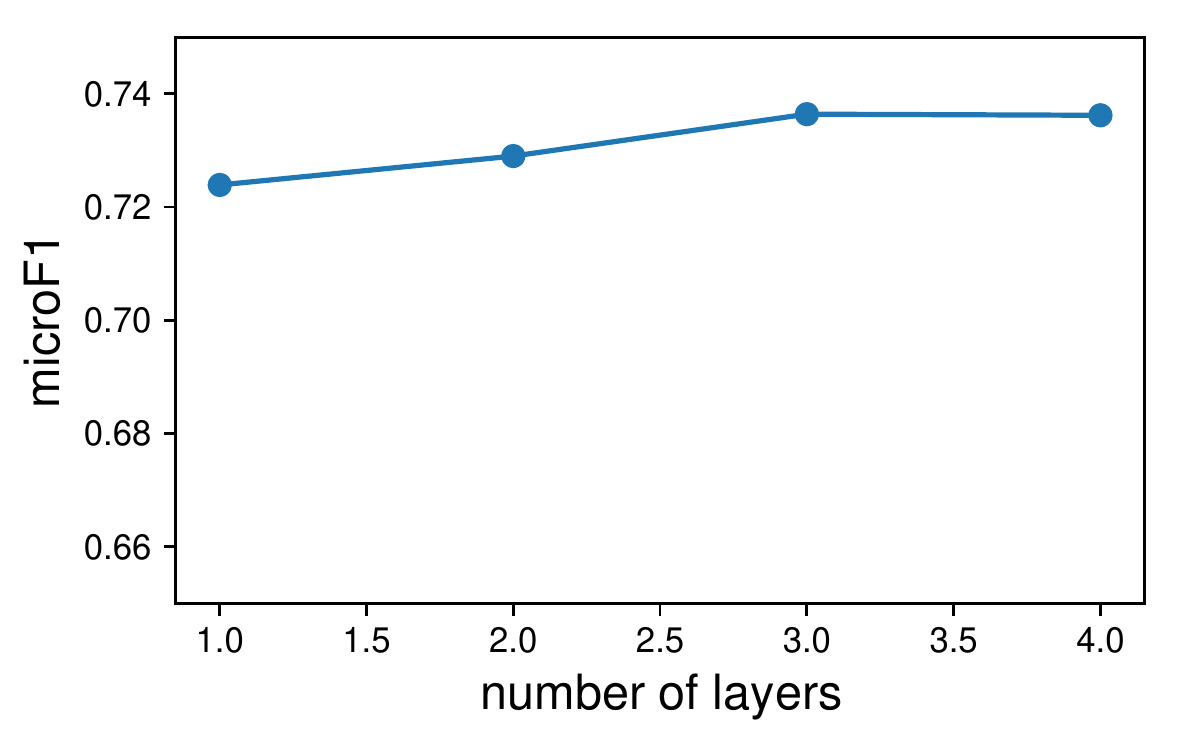}
    \end{subfigure}
    \begin{subfigure}[b]{0.31\textwidth}
        \centering
        \includegraphics[scale=0.33]{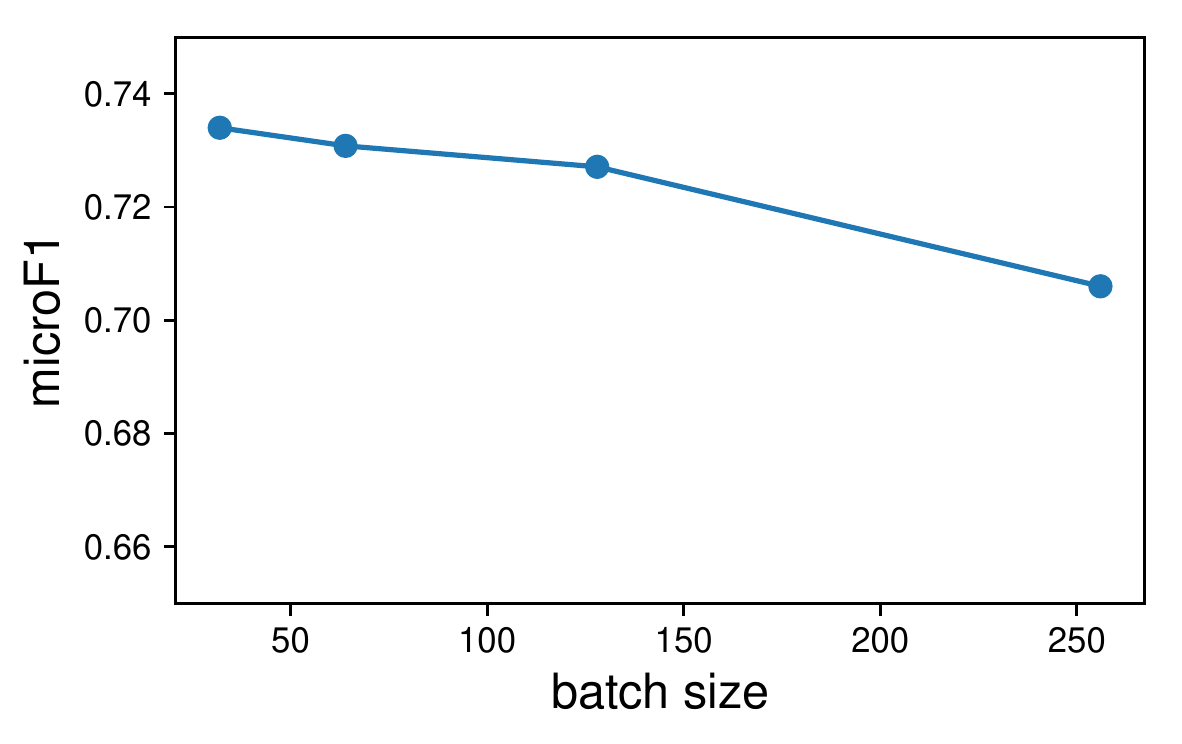}
    \end{subfigure}
    \begin{subfigure}[b]{0.31\textwidth}
        \centering
        \includegraphics[scale=0.33]{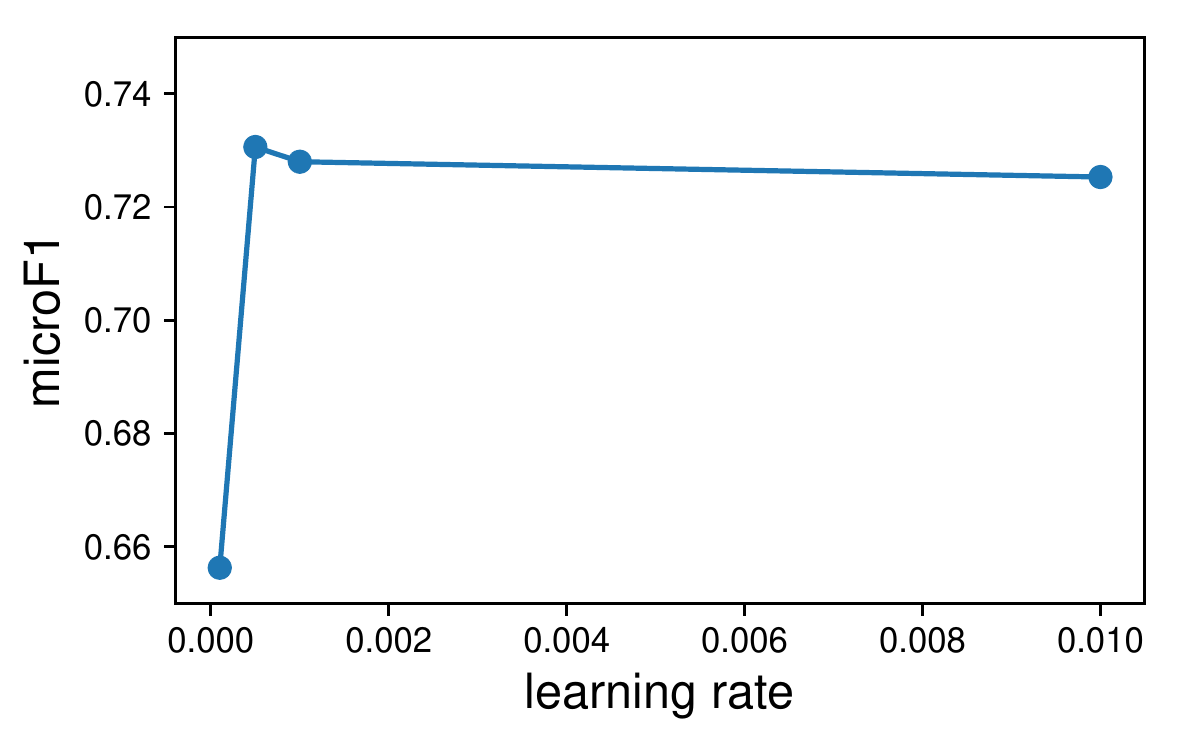}
    \end{subfigure}
    \begin{subfigure}[b]{0.31\textwidth}
        \centering
        \includegraphics[scale=0.33]{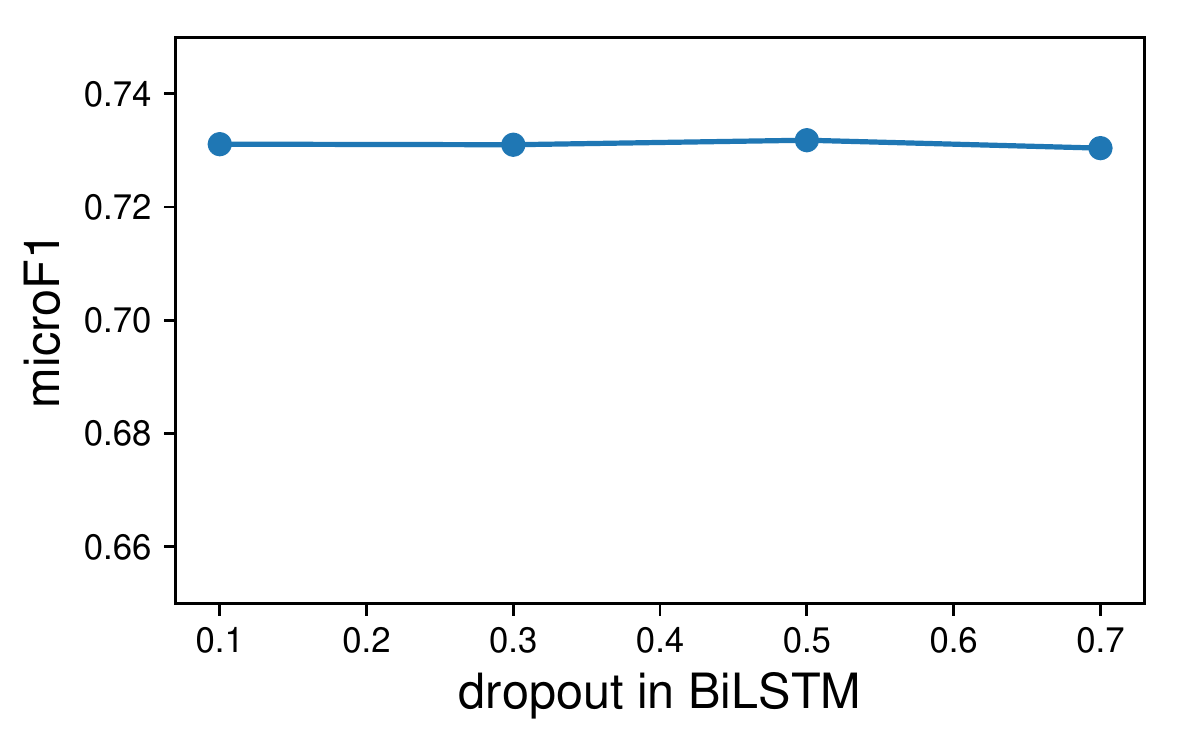}
    \end{subfigure}
    \begin{subfigure}[b]{0.31\textwidth}
        \centering
        \includegraphics[scale=0.33]{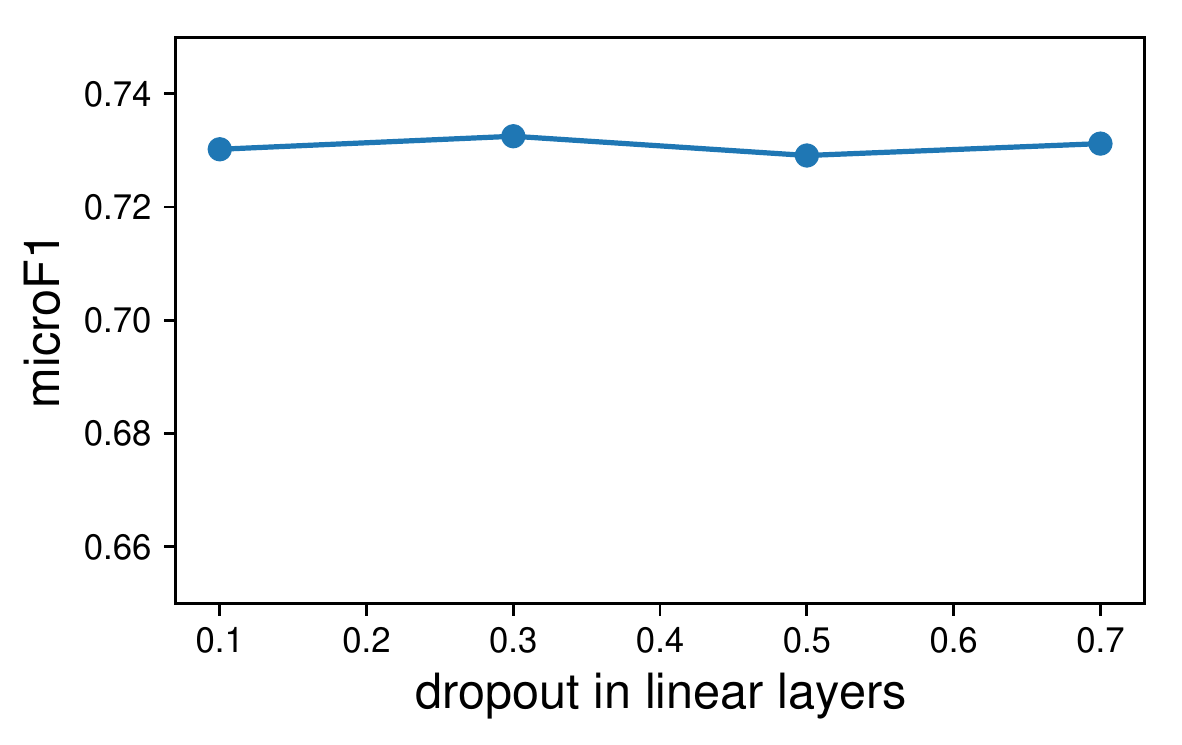}
    \end{subfigure}
    \caption{Sensitivity analysis on model hyper-parameters}
    \label{fig: sensitivity analysis}
\end{figure*}
In this section we explored different word and sentence representations and compared their performance on the test set. We also conducted sensitivity analysis for our model hyper-parameters. All results are averaged across 5 different seeds. It is worth noting that the settings with the best test scores in the analysis below are not exactly the same as our best system on the leaderboard since our best system is fine-tuned on the validation dataset, which do not guarantee to produce the best test results. 

We explored the original GloVe embedding trained on 27B tweet tokens\footnote{https://nlp.stanford.edu/projects/glove/}, pre-trained GloVe embedding\footnote{https://github.com/cbaziotis/datastories-semeval2017-task4}, our fine-tuned GloVe embedding, ELMo embedding and BERT embedding. The results are shown in Table 2. Fine-tuned GloVe embedding performs noticeably better than the original GloVe embedding and the pre-trained GloVe embedding. Surprisingly, contextualized embeddings such as ELMo and BERT perform worse than the original GloVe embedding. Possible reasons for their inferior performance are 1) they are fixed during training, which may hinder the overall optimization. 2) they have large embedding size, which can easily cause overfitting.

\begin{table}[t!]
\small
\begin{center}
\begin{tabular}{|c|c|}
\hline \bf Word Representation & \bf micro-F1 \\ \hline
Original Glove & 0.7250 \\
\hline
Pre-trained Glove & 0.7279 \\
\hline
Fine-tuned Glove & 0.7339 \\
\hline
ELMo & 0.6990\\
\hline
BERT & 0.6656\\
\hline
\end{tabular}
\end{center}
\caption{Micro-F1 score on the test set using different word representations}
\end{table}

\begin{table}[t!]
\small
\begin{center}
\begin{tabular}{|c|c|}
\hline \bf Sentence Representation & \bf micro-F1 \\ \hline
None & 0.7194 \\
\hline
InferSent (GloVe) & 0.7259 \\
\hline
InferSent (fastText) & 0.7277 \\
\hline
DeepMoji & 0.7299\\
\hline
DeepMoji + InferSent (GloVe) & 0.7298\\
\hline
DeepMoji + InferSent (fastText) & 0.7344\\
\hline
\end{tabular}
\end{center}
\caption{Micro-F1 score on the test set using different sentence representations}
\end{table}

We explored no sentence embedding, InferSent trained on GloVe, InferSent trained on fastText, and DeepMoji. The results are shown in Table 3. It is clear that sentence representations improved model performance significantly. In particular, DeepMoji achieves the best performance for single sentence representation. InferSent trained on fastText consistently performs better than InferSent trained on GloVe. In addition, concatenating two sentence representations together further improved model performance. 

We conducted sensitivity analysis on our model hyper-parameters: hidden size, number of layers in BiLSTM, batch size, learning rate, dropout in BiLSTM and dropout in linear layers. The results are depicted in Figure 2. Our model is relatively robust to hyper-parameters except for the learning rate. When learning rate is around 0.0001 or smaller, the model is unable to be trained effectively.

\section{Conclusion}
\label{conclusion}
In this paper we presented our model on the task of emotion detection in textual conversations in SemEval-2019. We explored different word and sentence representations in the RCNN model and achieved competitive results. Our result analysis indicate that both pre-trained word and sentence representations help improve the performance of RCNN. However, currently popular contextualized word representations such as ELMo and BERT produced inferior results. 

Future improvements can be made on the model architecture. In particular, simply concatenating three utterances into one sentence is not an information-preserving way to incorporate context information. We can design models that can handle a list of utterances and only classify the last utterance to optimize information flow.

\end{document}